\documentclass[11pt]{article}

\usepackage[preprint]{acl}

\usepackage{times}
\usepackage{latexsym}

\usepackage[T1]{fontenc}

\usepackage[utf8]{inputenc}

\usepackage{microtype}

\usepackage{inconsolata}

\usepackage{graphicx}

\usepackage{booktabs}
\usepackage{xcolor}
\usepackage{multirow}
\usepackage{amsmath}
\usepackage{cleveref}
\usepackage[most]{tcolorbox}
\AddToHook{cmd/appendix/before}{%
    \crefalias{section}{appendix}%
    \crefalias{subsection}{appendix}
}

\usepackage{tikz}
\usetikzlibrary{calc}

\usepackage{algorithm}
\usepackage{algpseudocode}

\usepackage{tabularx}
\usepackage{arydshln} %
\definecolor{errorred}{RGB}{180,40,40}
\newcommand{\err}[2]{\textcolor{errorred}{#1}\,\textcolor{errorred}{\scriptsize[\textsc{#2}]}}

\title{Reasoning over Grammar: Can Synthetic Linguistic Reasoning Traces Enhance Low-Resource Machine Translation?}

\author{
 \textbf{Renhao Pei\textsuperscript{1,2}},
 \textbf{Yihong Liu\textsuperscript{3,4}},
 \textbf{Sampo Pyysalo\textsuperscript{2}},
 \textbf{Hinrich Sch\"utze\textsuperscript{3,4}},
 \textbf{Shaoxiong Ji\textsuperscript{1,2}}
\\
\\
 \textsuperscript{1}ELLIS Institute Finland \space
 \textsuperscript{2}University of Turku\\
 \textsuperscript{3}Center for Information and Language Processing, LMU Munich\\
 \textsuperscript{4}Munich Center for Machine Learning (MCML)
\\
 \small{
\texttt{\{renpei,~sampo.pyysalo,~shaoxiong.ji\}@utu.fi}
\space
\texttt{yihong@cis.lmu.de}
 }
}

\begin{document}
\maketitle
\begin{abstract}

Large language models (LLMs) offer a promising approach to machine translation (MT) for extremely low-resource languages by incorporating linguistic resources through \emph{in-context learning}. 
However, LLMs often struggle to apply grammatical information effectively during translation. 
Inspired by recent progress in \emph{chain-of-thought reasoning}, we investigate whether low-resource MT can benefit from structured intermediate steps of linguistic analysis and grammatical reasoning. 
We propose a pipeline for automatically generating step-by-step linguistic reasoning traces from Universal Dependencies treebanks, dictionaries, and grammar-rule banks.
We evaluate these traces in three settings: in-context learning (ICL), supervised fine-tuning (SFT), and reinforcement fine-tuning (RFT), on \textbf{Xibe} and \textbf{Chintang} as test cases.
Our results show that linguistic reasoning traces are most effective as inference-time guidance: in ICL, reliable linguistic reasoning traces substantially improve translation performance across models, languages, and metrics. 
In contrast, using these traces as training data yields smaller and less consistent gains, as models learn the format but often generate error-prone content. 
These findings suggest that LLMs can leverage grammatical information for low-resource MT when given reliable linguistic analyses, while learning to generate such analyses remains a major bottleneck.\footnote{Our code and data are publicly available at: \href{https://olaresearch.org/LingReason}{https://olaresearch.org/LingReason}.}

\end{abstract}

\section{Introduction}

\begin{figure}[t]
  \centering
  \resizebox{\columnwidth}{!}{%
\begin{tikzpicture}[x=1cm,y=1cm,font=\scriptsize]
\definecolor{basegray}{HTML}{7A7A7A}    %
\definecolor{sftone}{HTML}{009E73}      %
\definecolor{sfttwo}{HTML}{0072B2}      %
\definecolor{sftthree}{HTML}{7B3294}    %
\definecolor{iclblue}{HTML}{D55E00}     %

\def\H{3.8}
\def\xA{0.7}
\def\xB{2.5}
\def\xC{4.3}
\def\xD{6.1}

\foreach \x/\name in {\xA/BLEU,\xB/chrF,\xC/SBERT,\xD/LLMaJ} {
  \draw[black!45] (\x,0) -- (\x,\H);
  \node[font=\scriptsize\bfseries,anchor=south] at (\x,\H+0.17) {\name};
}

\foreach \v in {0,2,4,6,8} {
  \pgfmathsetmacro{\y}{\H*\v/8}
  \draw[black!40] (\xA-0.07,\y) -- (\xA+0.07,\y);
  \node[anchor=east,text=black!65] at (\xA-0.11,\y) {\v};
}

\foreach \v in {0,10,20,30} {
  \pgfmathsetmacro{\y}{\H*\v/35}
  \draw[black!40] (\xB-0.07,\y) -- (\xB+0.07,\y);
  \node[anchor=east,text=black!65] at (\xB-0.11,\y) {\v};
}

\foreach \v in {0,20,40,60} {
  \pgfmathsetmacro{\y}{\H*\v/70}
  \draw[black!40] (\xC-0.07,\y) -- (\xC+0.07,\y);
  \node[anchor=east,text=black!65] at (\xC-0.11,\y) {\v};
}

\foreach \v in {0,10,20,30,40,50} {
  \pgfmathsetmacro{\y}{\H*\v/50}
  \draw[black!40] (\xD-0.07,\y) -- (\xD+0.07,\y);
  \node[anchor=east,text=black!65] at (\xD-0.11,\y) {\v};
}

\draw[basegray,dashed,line width=0.9pt]
  (\xA,{3.28/8*\H}) --
  (\xB,{20.84/35*\H}) --
  (\xC,{46.23/70*\H}) --
  (\xD,{24.43/50*\H});
\foreach \x/\y in {
  \xA/{3.28/8*\H},\xB/{20.84/35*\H},
  \xC/{46.23/70*\H},\xD/{24.43/50*\H}} {
  \fill[basegray] (\x,\y) circle (1.25pt);
}

\draw[sftone,line width=0.85pt]
  (\xA,{4.02/8*\H}) --
  (\xB,{23.55/35*\H}) --
  (\xC,{52.47/70*\H}) --
  (\xD,{28.69/50*\H});
\foreach \x/\y in {
  \xA/{4.02/8*\H},\xB/{23.55/35*\H},
  \xC/{52.47/70*\H},\xD/{28.69/50*\H}} {
  \fill[sftone] (\x,\y) circle (1.35pt);
}

\draw[sfttwo,line width=0.85pt]
  (\xA,{4.13/8*\H}) --
  (\xB,{25.13/35*\H}) --
  (\xC,{53.35/70*\H}) --
  (\xD,{26.44/50*\H});
\foreach \x/\y in {
  \xA/{4.13/8*\H},\xB/{25.13/35*\H},
  \xC/{53.35/70*\H},\xD/{26.44/50*\H}} {
  \fill[sfttwo] (\x,\y) circle (1.35pt);
}

\draw[sftthree,line width=0.85pt]
  (\xA,{4.00/8*\H}) --
  (\xB,{25.51/35*\H}) --
  (\xC,{53.56/70*\H}) --
  (\xD,{25.55/50*\H});
\foreach \x/\y in {
  \xA/{4.00/8*\H},\xB/{25.51/35*\H},
  \xC/{53.56/70*\H},\xD/{25.55/50*\H}} {
  \fill[sftthree] (\x,\y) circle (1.35pt);
}

\draw[iclblue,line width=1.45pt]
  (\xA,{7.02/8*\H}) --
  (\xB,{31.19/35*\H}) --
  (\xC,{64.80/70*\H}) --
  (\xD,{45.55/50*\H});
\foreach \x/\y in {
  \xA/{7.02/8*\H},\xB/{31.19/35*\H},
  \xC/{64.80/70*\H},\xD/{45.55/50*\H}} {
  \fill[iclblue] (\x,\y) circle (1.8pt);
}

\begin{scope}[shift={(0.55,-0.58)}, font=\scriptsize]

  \draw[basegray,dashed,line width=0.9pt] (0,0) -- (0.38,0);
  \fill[basegray] (0.19,0) circle (1.25pt);
  \node[anchor=west] at (0.48,0) {Baseline};

  \draw[sfttwo,line width=0.85pt] (0,-0.34) -- (0.38,-0.34);
  \fill[sfttwo] (0.19,-0.34) circle (1.35pt);
  \node[anchor=west] at (0.48,-0.34) {SFT w/ reasoning};

  \draw[iclblue,line width=1.45pt] (0,-0.68) -- (0.38,-0.68);
  \fill[iclblue] (0.19,-0.68) circle (1.8pt);
  \node[anchor=west] at (0.48,-0.68) {ICL};

  \draw[sftone,line width=0.85pt] (3.10,0) -- (3.48,0);
  \fill[sftone] (3.29,0) circle (1.35pt);
  \node[anchor=west] at (3.58,0) {SFT w/o reasoning};

  \draw[sftthree,line width=0.85pt] (3.10,-0.34) -- (3.48,-0.34);
  \fill[sftthree] (3.29,-0.34) circle (1.35pt);
  \node[anchor=west] at (3.58,-0.34) {SFT w/ reasoning + RFT};

\end{scope}
\end{tikzpicture}
}
\caption{Comparison of Qwen3-8B translation performance on Chintang 
across the baseline (in-context MT without reasoning), SFT, RFT, and ICL settings. 
ICL clearly outperforms the training-based settings on all four metrics, suggesting that linguistic reasoning traces are most useful as reliable inference-time guidance rather than as training supervision.}
  \label{fig:parallel-overview}
\end{figure}
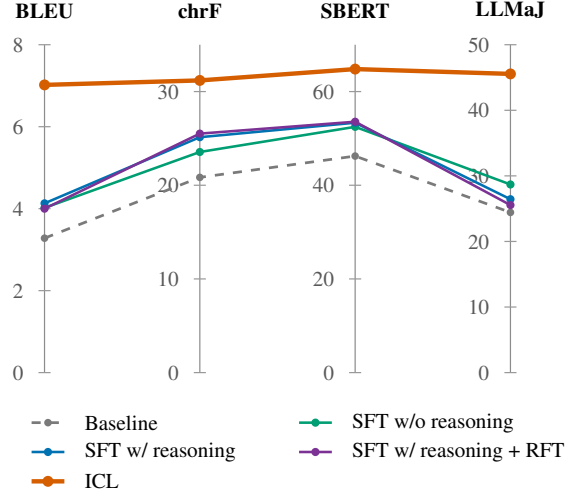

Only a small fraction of the world's more than 7,000 languages have sufficient parallel data for training dedicated machine translation (MT) systems, and for many low-resource languages, such data are scarce or entirely unavailable \citep{bapna2022buildingmachinetranslationsystems}.
At the same time, many of these languages are well documented through linguistic resources such as dictionaries, grammar books, and annotated treebanks \citep{nordhoff2011glottolog}. 

To bridge the gap between scarce parallel data and comparatively abundant linguistic resources, recent work has explored using large language models (LLMs) for in-context MT, where dictionaries, grammar descriptions, or example sentences are incorporated into the prompt alongside the sentence to be translated \citep{tanzer2024a,zhang-etal-2024-hire, hus-anastasopoulos-2024-back, zhang-etal-2024-teaching, pei-etal-2025-understanding}.

However, making effective use of grammatical information remains challenging. 
Grammatical rules that describe morphemes, syntactic constructions, and compositional structures are crucial for understanding low-resource languages, and human translators often rely on such information through explicit linguistic analysis \citep{neacșu2024linguistics}. 
Yet prior work has shown that, while LLMs can often benefit from lexical information, they struggle to reason over grammatical descriptions during in-context MT \citep{aycock2025can, pei-etal-2025-understanding}. 
This limitation suggests that simply placing grammar rules in the prompt may not be sufficient: models may need a more structured procedure that guides them through how grammatical information should be applied during translation.

Motivated by recent progress in chain-of-thought (CoT) reasoning, where explicit intermediate steps have improved performance on complex tasks such as mathematics and puzzle solving \citep{wei2022cot,ahn-etal-2024-large, giadikiaroglou-etal-2024-puzzle}, we ask whether \emph{low-resource MT can benefit from structured linguistic reasoning}. 
More specifically, instead of treating translation as a direct sequence-to-sequence mapping, we investigate whether LLMs can translate more effectively when guided to decompose a sentence, analyze its lexical and morphosyntactic structure, apply relevant grammar rules, and compose intermediate phrasal meanings into a final translation. 
Since no comparable dataset of linguistic reasoning traces exists for this type of translation task, we first propose a pipeline for automatically generating step-by-step linguistic reasoning traces from Universal Dependencies (UD) treebanks, dictionaries, and modular grammar-rule banks.

We evaluate the generated reasoning traces in three experimental settings: in-context learning (ICL), supervised fine-tuning (SFT), and reinforcement fine-tuning (RFT).
For each setting, we compare against a corresponding baseline without reasoning traces. 
As illustrated in \Cref{fig:parallel-overview}, our results show that linguistic reasoning traces are most effective when used as inference-time guidance: in the ICL setting, reliable sentence-specific traces substantially improve translation performance over the baseline and outperform the training-based settings. 
In contrast, when the same traces are used as training data, SFT and RFT yield smaller and less consistent gains, suggesting that models can benefit from reliable linguistic analyses but still struggle to generate such analyses accurately by themselves.

The contributions of this work are as follows:

\textbf{(i) We develop a pipeline for automatically generating step-by-step linguistic reasoning traces.}
The pipeline incorporates UD treebanks, dictionaries, and modularized grammar rules. 
To the best of our knowledge, this is the first framework for constructing such reasoning traces for the MT of extremely low-resource languages.

\textbf{(ii) We evaluate whether LLMs can reason over grammar through both prompting and fine-tuning.}
Our experiments cover three settings: ICL, SFT, and RFT. 
While prior work has mainly focused on prompting-based in-context MT, we further examine whether linguistic reasoning traces can serve as supervision for fine-tuning.

 \textbf{(iii) We identify where linguistic reasoning traces help most.}
Our results show that structured linguistic reasoning traces are currently more effective as inference-time guidance than as training supervision. This suggests that LLMs can benefit from grammatical information when given reliable analyses in the context, but still struggle to generate such analyses on their own.

\section{Related Work}

\paragraph{In-context MT for Low-Resource Languages.}

Since \citet{tanzer2024a} introduced Machine Translation from One Book (MTOB), various studies have investigated incorporating linguistic resources such as dictionary entries, parallel examples, interlinear glosses and grammar books into prompts, and leveraging LLMs' in-context learning abilities for low-resource MT~\citep{zhang-etal-2024-hire, hus-anastasopoulos-2024-back, zhang-etal-2024-teaching, pei-etal-2025-understanding, ramos-etal-2025-grammamt}.

While adding dictionary entries consistently improves performance, \citet{aycock2025can} point out that the gains from using grammar books come only from the parallel example sentences in them, and LLMs are unable to effectively use grammatical explanations to improve translation.
Similar findings are reported by \citet{pei-etal-2025-understanding}, showing that adding grammatical information does not improve in-context MT, and the attempt to address this with CoT prompting only further degrades performance.

To disentangle the retrieval and application of grammatical information, \citet{zhang-etal-2025-read} construct a dataset of grammar rules paired with relevant example sentences. Their findings indicate that grammar rule retrieval is a bottleneck, and LLMs also struggle with complex grammar rules.

\citet{purushothama2026syntaxrosettastoneuniversal} incorporate UD treebanks into the context to improve translation.
Their results indicate the potential usefulness of syntactic information with modest gains, while the underlying tree structure is not exploited explicitly.

In contrast, we leverage the UD tree structure directly to generate step-by-step reasoning traces that mirror the syntactic composition of the sentence.

\paragraph{LLM reasoning for MT.}

Recent work has explored various ways of eliciting translation-oriented reasoning from LLMs.
\citet{briakou-etal-2024-translating} propose a multi-turn translation of pre-translation research, drafting, refinement, and proofreading, whereas \citet{wu-etal-2025-please} investigate iterative self-refinement and show that simply prompting models to translate again can outperform more elaborate methods. \citet{rajaee2026unlockingreasoningcapabilitymachine} further propose a multi-stage framework including initial drafting, adequacy enhancement, fluency refinement, and selective revision. \citet{he2025r1t1fullyincentivizingtranslation} introduce human-aligned CoT templates and RL to elicit inference-time reasoning for MT, while \citet{zheng2025hunyuanmttechnicalreport} train Hunyuan-MT through a multilingual translation pipeline with SFT and RL.

However, these works are primarily aimed at further improving MT performance for relatively high-resource languages, by decomposing translation into stages such as drafting and refinement. In contrast, our approach targets extremely low-resource languages, where basic translation adequacy remains challenging. Our step-by-step reasoning therefore focuses on linguistic reasoning over grammatical information to help recover the basic semantics of the source sentence, rather than polishing an already plausible translation.

\section{Languages, Data and General Setup}\label{setup}

\paragraph{Languages.} 

Xibe (ISO 639-3: \texttt{sjo}) is a Tungusic language spoken in Northwest China, with around 30,000 native speakers.\footnote{Xibe and the historically prominent Manchu language share an almost identical literary language, so that Manchu dictionaries and grammar books can also be used as supplementary resources for Xibe.} It exemplifies the setting in which external linguistic resources, including dictionaries and grammar books, are incorporated alongside UD treebanks.

Chintang (ISO 639-3: \texttt{ctn}) is a Sino-Tibetan language spoken in Nepal, with around 5,000 speakers. It exemplifies a setting that relies only on UD data.
These two languages therefore represent two distinct resource scenarios.

The translation direction in all our experiments is always from low-resource language to English.\footnote{Current LLMs struggle to generate fluent, grammatical text in these extremely low-resource languages. By focusing on translation from low-resource languages into English, we can isolate translation adequacy from target-language generation quality.}
Although our evaluation is limited to two languages due to resource constraints, they represent distinct resource settings, and the consistent results suggest that the approach is not specific to a single language or resource type.

\paragraph{UD treebanks.}
UD is a cross-linguistic framework for morphosyntactic annotation based on dependency grammar, where sentence structure is represented as head--dependent relations between words, and the relation between them is expressed by a dependency label indicating the grammatical function of the dependent \citep{de-marneffe-etal-2021-universal}. UD annotations include word forms, lemmas, part-of-speech (POS) tags, dependency relations, and morphological features, together with optional information such as word-level glosses, transliterations\footnote{Xibe uses non-Latin scripts while its UD includes Latin transliterations, which are used throughout our experiments.}, and sentence-level translations\footnote{The UD treebanks of both Xibe and Chintang provide sentence-level English translations, which are used as parallel data for our MT experiments.}.

In our experiments, a maximum sentence-length filter of 30 words is applied, which keeps 979 of 1,200 trees for Xibe treebank and 2,289 of 2,289 trees for Chintang.

\paragraph{Dictionaries.}

The Xibe dictionary data are drawn from \citet{norman2000sibe} and the online dictionary \textit{Mini Buleku} \citep{kodner2021minibuleku}. 
\footnote{Licensed under a Creative Commons Attribution-NonCommercial-NoDerivatives 4.0 International License.}
The dictionary data are further supplemented with the Manchu dictionary of \citet{norman2020comprehensive}\footnote{Accessed via \url{https://buleku.org/home}; used with permission from the author.} and explanations of Manchu suffixes 
based on \citet{clark1980manchu}. 
The Xibe dictionary entries always take precedence over the Manchu entries.

For Chintang,
the UD treebanks natively include English glosses for each lemma as part of their annotations, which we use to construct dictionaries. 
Inflectional or morphological annotations are removed from lexical entries, while grammatical annotations are retained for purely grammatical morphemes. Different glosses attested for the same lemma are merged into a single polysemous dictionary entry.

\paragraph{Grammar rules.}
The grammar resources 
are organized as collections of separate grammar rules. Each rule consists of a short textual explanation of a particular grammatical phenomenon, paired with a UD-based trigger, such as a specific dependency relation, feature--value pair, POS tag, or, where useful, a combination of these features \footnote{A rule is invoked when its trigger is encountered at a given composition step during the reasoning trace generation, as illustrated in \Cref{fig:example}.}.

For Xibe, the grammar rules are primarily based on manually selected excerpts from \citet{zhou-etal-2020-universal} and \citet{gorelova2002manchu}, further supplemented with explanations from the UD language documentation pages.
For Chintang, %
the rules are derived by matching UD features with the corresponding explanations in its highly detailed UD documentation pages.
The final grammar-rule set contains 77 rules for Xibe and 82 rules for Chintang.

These modularized grammar rules serve as a grammatical knowledge bank that can be automatically matched against UD-parsed sentences and incorporated into the generated reasoning traces.

\paragraph{Models.}\label{para:models}
We conduct our experiments on two model families with varying sizes: Qwen3 \citep{yang2025qwen3technicalreport}, including 4B, 8B, and 14B models, and Gemma 4 \citep{google_deepmind_gemma4_2026}, including E2B, E4B, and 31B models. Based on our pilot experiments, we use only instruction-tuned models, as they outperform their base-model counterparts.

For the 4B Qwen3 model, we use the thinking-only variant Qwen3-4B-Thinking-2507\footnote{Shorthanded as Qwen3-4B-Thinking in tables.}, as it outperforms the non-thinking variant Qwen3-4B-Instruct-2507.
The other models in our experiments all support seamless switching between thinking and non-thinking modes.
Using models with reasoning capabilities allows us to take advantage of their general ability of step-by-step reasoning.

For all experiments, we follow the recommended decoding hyperparameters from the corresponding model cards. Details are provided in Appendix~\ref{sec:implementation_details}.

\paragraph{Evaluation metrics.}\label{sec:metrics}
To measure the translation quality, we use BLEU \citep{papineni-etal-2002-bleu} and chrF \citep{popovic-2015-chrf} to measure word-level and character-level n-gram overlap, as implemented by SacreBLEU \citep{post-2018-call}.\footnote{BLEU signature: nrefs:1|case:lc|eff:no|tok:13a|\newline smooth:exp|version:2.6.0\newline
chrF signature: nrefs:1|case:mixed|eff:yes|nc:6|nw:0|\newline space:no|version:2.6.0
}
We also report SBERT \citep{reimers-2019-sentence-bert}, an embedding-based metric that assesses the semantic relatedness between a translation and a reference sentence.\footnote{SBERT score is computed using the \texttt{all-MiniLM-L6-v2} sentence-transformer model and the score is multiplied by 100 for a uniform magnitude across metrics.}

Additionally, we also employ LLM-as-a-judge (LLMaJ, \citealt{chiang-lee-2023-large, NEURIPS2023_91f18a12}) as an evaluation method.
The judge model (Gemini 3.1 Flash-Lite) is asked to rate the generated translation on a scale from 0 to 100, based on the gold-standard reference translation.
The LLMaJ prompt template is adapted from the WMT25 template \citep{kocmi-etal-2025-findings-wmt25} and the human evaluation instructions of \citet{pei-etal-2025-understanding}. It focuses on adequacy rather than fluency, since translations generated in the in-context MT are almost always fluent and grammatical in English. The template is provided in \Cref{sec:prompt_templates_llm_judge}.

\section{Generation of Linguistic Reasoning Traces}

\begin{figure}
    \centering
    \setlength{\belowcaptionskip}{-0.5cm}
    \includegraphics[width=0.49\textwidth]{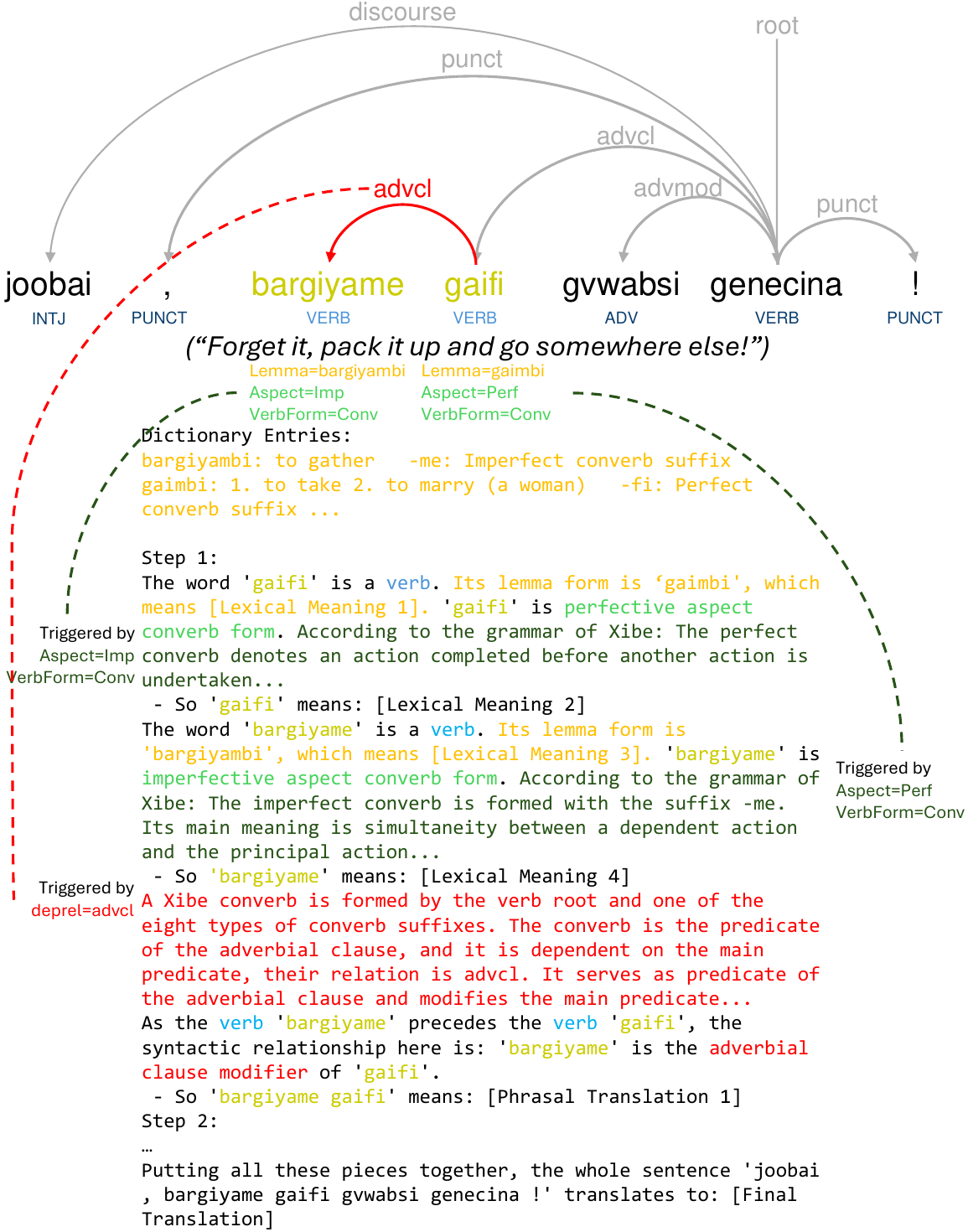}
    \caption{An illustration of the generated reasoning trace of a Xibe UD tree. UD tokens and tags are color-matched with their corresponding text in the generated reasoning trace. Placeholders are not yet filled in.}
    \label{fig:example}
\end{figure}

Utilizing the available linguistic resources from UD treebanks, dictionaries, and grammar rules, we design a pipeline for generating step-by-step reasoning traces that start from the lexical and morphological meanings of individual words, through steps of intermediate phrasal translations, and finally reach the full sentential translation. These traces incorporate language-specific grammar rules, syntactic relations between words, and partial phrasal meanings, illustrating the progressive procedure of composing smaller linguistic units into larger ones through grammatical analysis. Figure~\ref{fig:example} shows an illustration of the generated reasoning trace.

\paragraph{Traversing of the UD tree as the order of reasoning steps.}
In the step-by-step linguistic reasoning, each step corresponds to combining a head with its dependent(s). The steps are ordered bottom-up according to post-order traversal of the UD tree: all child subtrees are traversed before their parent, so that a step for a smaller subtree always appears before any step in which that subtree is incorporated into a larger subtree.

In this order, each non-leaf node with its immediate children is converted into one composition step. Each step therefore centers on a single head and its relation with the dependent(s). When a head has multiple dependents, they are processed in the order of ascending index, i.e., left-to-right surface order.

This composition procedure is well aligned with the principle of compositionality in semantics, that the meaning of a complex expression is derived from the meanings of its parts and from the way those parts are combined.

\paragraph{Verbalizing each step using linguistic resources.}
Before the reasoning trace itself, dictionary entries for all words occurring in the sentence are listed in the prompt.

At each reasoning step, each token, whether a head or dependent, is first described by verbalizing its POS tag, lemma, and morphological features. Relevant grammar rules concerning specific morphemes are inserted when triggered by the token's features. A \texttt{[Lexical Meaning]} placeholder is then inserted after the word-level explanation.

Each syntactic relation is then verbalized by specifying the linear order of the head and dependent, their POS tags, and the UD dependency relation between them. When triggered by the dependency relation, relevant grammar rules concerning the syntactic structure are inserted before the verbalization of the relation, creating a reasoning flow in which the grammar rule leads the identification of the syntactic relation. A \texttt{[Phrasal Meaning]} placeholder is inserted after the explanation of the syntactic relation between the head and the dependent.

This process converts UD trees into step-by-step reasoning traces that contain the relevant lexical and grammatical information, with some placeholders not yet filled in. These reasoning traces with placeholders are used as in-context guidance for the LLM in our in-context MT experiments.

\paragraph{LLM filling in placeholders.}
For use as training data in SFT and RFT, the placeholders in the reasoning traces are further filled in by an LLM (Gemini 3.1 Flash-Lite Preview). For Xibe (\texttt{sjo}), both lexical meanings and phrasal translations are filled in by the LLM. For Chintang,
lexical meanings are already available from UD annotations, so only the phrasal translation placeholders need to be filled.

To fill in the placeholders, the LLM is provided with dictionary entries and the final gold sentence translation as contextual cues. The task is therefore to select the appropriate sense for polysemous words based on their meaning in the sentence-level translation, and to derive intermediate phrasal translations when both word-level and sentence-level translations are available. The prompt template is provided in \Cref{sec:prompt_templates_llm_fill_in_placeholder}.
The structure of the reasoning trace is therefore defined by our template and does not mirror the LLM's own reasoning process.

The filled-in lexical and phrasal translations can subsequently serve as structured intermediate supervision signals, which are used for the process reward described in \Cref{sec:reward_function}.

\section{In-Context Learning Experiment}

\subsection{Setup}

We conduct the in-context learning experiment with two prompting variants
and evaluate their performance on the test split (15\% of the full treebank) 
as described in \Cref{sec:metrics}. 
The implementation details are provided in \Cref{sec:icl-setting}, and the prompt templates are provided in \Cref{sec:prompt_templates_in_context_mt}.

The \textbf{baseline} prompt includes the relevant dictionary entries and all grammar rules triggered by the sentence's UD annotation, i.e., the same rules used in the reasoning trace.\footnote{The main comparison here focuses on whether structuring the same grammatical information into explicit reasoning improves translation. The baseline without any grammatical information can be found in \Cref{sec:dictionary_only_baseline}.} Thus, it contains the same linguistic information as the reasoning variant, but without organizing it into step-by-step reasoning.

The \textbf{\textit{+reasoning}} prompt includes the UD-derived linguistic reasoning traces with placeholders, instead of the flat list of grammar rules. Each trace provides an explicit, sentence-specific analytic path for translation. 
At inference time, the LLM is instructed to resolve these placeholders one by one before outputting the final translation.

We additionally include a simple rule-based \textbf{word-by-word} baseline to contextualize the translation performance: each source sentence is translated word by word,\footnote{For polysemous words, the first listed English sense is selected. Words not found in the dictionary are copied unchanged.} using the same lexical resources provided to the models in the ICL setting. 

\subsection{Results and Analysis}

\begin{table*}[t]
\centering
\small
\setlength{\tabcolsep}{4pt}
\renewcommand{\arraystretch}{1.12}
\newcommand{\gain}[2]{\textbf{#1} {\scriptsize\textcolor{teal}{(+#2)}}}
\newcommand{\loss}[2]{#1 {\scriptsize\textcolor{gray}{(#2)}}}
\resizebox{\textwidth}{!}{%
\begin{tabular}{@{}lcccc:cccc@{}}
\toprule
\multirow{2}{*}{Model}
& \multicolumn{4}{c:}{Xibe (\texttt{sjo})}
& \multicolumn{4}{c}{Chintang (\texttt{ctn})} \\
\cmidrule(lr){2-5}\cmidrule(lr){6-9}
& BLEU & chrF & SBERT & LLMaJ & BLEU & chrF & SBERT & LLMaJ \\
\midrule

gemma-4-E2B-it
& 0.76 & 22.15 & 47.57 & 42.87
& 3.79 & 22.51 & 49.20 & 31.77 \\
\quad +reasoning
& \loss{0.58}{-0.18} & \loss{17.49}{-4.66} & \gain{51.30}{3.73} & \loss{39.48}{-3.39}
& \loss{3.37}{-0.42} & \gain{24.08}{1.57} & \gain{51.26}{2.06} & \gain{41.54}{9.77} \\
\cdashline{1-9}

gemma-4-E4B-it
& 0.67 & 16.05 & 49.35 & 40.01
& 1.39 & 17.86 & 53.72 & 31.48 \\
\quad +reasoning
& \gain{1.88}{1.21} & \gain{22.70}{6.65} & \gain{54.49}{5.14} & \gain{45.38}{5.37}
& \gain{6.96}{5.57} & \gain{29.75}{11.89} & \gain{65.04}{11.32} & \gain{49.99}{18.51} \\
\cdashline{1-9}

gemma-4-31B-it
& 9.84 & 35.85 & 62.31 & 59.69
& 9.96 & 31.61 & 63.13 & 43.24 \\
\quad +reasoning
& \gain{10.81}{0.97} & \gain{37.36}{1.51} & \gain{64.28}{1.97} & \gain{63.29}{3.60}
& \gain{12.59}{2.63} & \gain{36.53}{4.92} & \gain{73.46}{10.33} & \gain{65.14}{21.90} \\
\cdashline{1-9}

Qwen3-4B-Thinking
& 0.18 & 6.71 & 35.72 & 28.60
& 0.22 & 5.86 & 44.03 & 25.14 \\
\quad +reasoning
& \gain{1.21}{1.03} & \gain{17.54}{10.83} & \gain{53.21}{17.49} & \gain{41.82}{13.22}
& \gain{1.11}{0.89} & \gain{13.12}{7.26} & \gain{63.77}{19.74} & \gain{48.56}{23.42} \\
\cdashline{1-9}

Qwen3-8B
& 2.08 & 26.69 & 48.43 & 35.97
& 3.28 & 20.84 & 46.23 & 24.43 \\
\quad +reasoning
& \gain{2.74}{0.66} & \gain{27.79}{1.10} & \gain{55.57}{7.14} & \gain{43.23}{7.26}
& \gain{7.02}{3.74} & \gain{31.19}{10.35} & \gain{64.80}{18.57} & \gain{45.55}{21.12} \\
\cdashline{1-9}

Qwen3-14B
& 3.11 & 28.99 & 52.79 & 43.56
& 5.25 & 26.50 & 56.10 & 32.91 \\
\quad +reasoning
& \gain{4.70}{1.59} & \gain{30.83}{1.84} & \gain{57.09}{4.30} & \gain{48.74}{5.18}
& \gain{10.33}{5.08} & \gain{34.07}{7.57} & \gain{67.07}{10.97} & \gain{51.98}{19.07} \\
\midrule

word-by-word
& 0.30 & 20.95 & 31.82 & 34.03
& 1.09 & 15.82 & 37.42 & 35.17 \\

\bottomrule
\end{tabular}%
}
\caption{ICL performance on \texttt{sjo} and \texttt{ctn}. Gains from adding reasoning traces with placeholders are shown in parentheses, and \textbf{bold} indicates scores higher than the baseline. Adding reasoning traces substantially improves performance across languages, metrics, and models, with the exception of gemma-4-E2B-it. A simple word-by-word translation baseline is included for comparison.}
\label{tab:icl-results}
\end{table*}

\textbf{Adding reasoning traces substantially improves in-context MT performance for most models.}
As shown in Table~\ref{tab:icl-results}, adding reasoning traces with placeholders yields gains across languages, metrics, and models. The gains are especially consistent for SBERT, suggesting that reasoning-guided prompts help models produce translations that are semantically closer to the references.

Improvements are particularly large for \texttt{ctn}, where adding reasoning traces yields substantial gains: up to +5.57 BLEU and +11.89 chrF on gemma-4-E4B-it, and up to +19.74 SBERT and +23.42 LLMaJ on Qwen3-4B-Thinking-2507. For \texttt{sjo}, the improvements are more moderate but still mostly positive.

\textbf{Benefits from reasoning traces are less consistent for the smallest model.}
The only exception is gemma-4-E2B-it, for which BLEU decreases while SBERT increases, and chrF and LLMaJ show mixed results across languages. This may be due to the model's smaller capacity and lower baseline performance, making its outputs more susceptible to noise.

Overall, the ICL results show that structured linguistic reasoning provides useful in-context guidance and can substantially improve translation performance. It is worth noting that, even when provided with a step-by-step reasoning trace in the \textbf{\textit{+reasoning}} setting, the model may still fill some placeholders incorrectly. These intermediate errors are associated with lower final translation quality. A detailed analysis can be found in \Cref{sec:intermediate_correctness}.

Compared with the word-by-word baseline, the +reasoning ICL settings achieve substantially higher scores across nearly all languages, metrics, and models, with a few exceptions for the smallest models, gemma-4-E2B-it and Qwen3-4B-Thinking. This indicates that \textbf{the ICL performance cannot be attributed to dictionary lookup alone}; rather, the models benefit from their in-context learning capability to combine lexical information with sentence-level reasoning over morphology, syntax, and compositional meaning.

\section{Supervised Fine-Tuning Experiment}

Although adding linguistic reasoning in the ICL setting yields strong gains without any additional training, this approach is not readily applicable to new sentences without accurate UD parses.
To examine whether the reasoning traces can be used to train models to generalize linguistic reasoning to unseen data, we conduct a supervised fine-tuning (SFT) experiment.

\subsection{Dataset}
We construct a fine-tuning dataset for in-context MT using the completed reasoning traces, i.e., traces in which all placeholders have been filled in. Each dataset instance consists of a prompt and an answer; the SFT objective is therefore to train the model to generate the answer given the prompt.

The \textbf{prompt} follows the baseline prompt template as in \Cref{sec:prompt_templates_in_context_mt}, containing the MT task instructions, relevant dictionary entries for each word in the source-language sentence, the grammar rules triggered by the UD tags, and the source-language sentence to be translated. 
The \textbf{answer} contains the generated reasoning trace enclosed in \texttt{<think>}...\texttt{</think>}, followed by the final English translation enclosed in \texttt{<answer>}...\texttt{</answer>}. 

The dataset is built using the whole UD treebanks, and is split into 80\% training, 5\% validation, and 15\% test sets. The validation set is used to select the best checkpoints, and the test set is used to compute the final scores for reporting, which contains the same sentences as those used in the previous ICL experiment.

\subsection{Setup}

This experiment compares two SFT settings and evaluates how they perform relative to the models before fine-tuning. Due to limited computing resources, we exclude the 14B and 31B models. Implementation details are provided in \Cref{sec:sft-setting}.

For the \textit{\textbf{SFT without reasoning}} setting, we fine-tune the model on prompts paired only with final translations, excluding the reasoning traces enclosed in the \texttt{<think>} block.
For the \textit{\textbf{SFT with reasoning}} setting, we fine-tune the model on full training answers containing both reasoning traces and final translations. The model is therefore trained to first generate a reasoning trace and then produce the final translation.

Final translations are extracted from the \texttt{<answer>} block of the fine-tuned models' outputs and evaluated on the same test set. The difference between the two SFT settings therefore reflects the effect of including reasoning traces in the SFT training data.

\subsection{Results and Analysis}

\begin{table*}[t]
\centering
\small
\setlength{\tabcolsep}{4pt}
\renewcommand{\arraystretch}{1.12}
\newcommand{\gain}[2]{#1 {\scriptsize\textcolor{teal}{(+#2)}}}
\newcommand{\loss}[2]{#1 {\scriptsize\textcolor{gray}{(#2)}}}
\newcommand{\bgain}[2]{\textbf{#1} {\scriptsize\textcolor{teal}{(+#2)}}}
\newcommand{\bloss}[2]{\textbf{#1} {\scriptsize\textcolor{gray}{(#2)}}}
\resizebox{\textwidth}{!}{%
\begin{tabular}{@{}lcccc:cccc@{}}
\toprule
\multirow{2}{*}{Model}
& \multicolumn{4}{c:}{Xibe (\texttt{sjo})}
& \multicolumn{4}{c}{Chintang (\texttt{ctn})} \\
\cmidrule(lr){2-5}\cmidrule(lr){6-9}
& BLEU & chrF & SBERT & LLMaJ & BLEU & chrF & SBERT & LLMaJ \\
\midrule

gemma-4-E2B-it
& 0.76 & 22.15 & 47.57 & 42.87
& 3.79 & 22.51 & 49.20 & 31.77 \\
\quad +SFT w/o reasoning
& \gain{1.10}{0.34} & \gain{22.84}{0.69} & \loss{42.50}{-5.07} & \loss{34.77}{-8.10}
& \loss{1.99}{-1.80} & \loss{20.25}{-2.26} & \loss{39.86}{-9.34} & \loss{18.85}{-12.92} \\
\quad +SFT w/ reasoning
& \bgain{1.24}{0.48} & \loss{20.71}{-1.44} & \loss{37.44}{-10.13} & \loss{25.64}{-17.23}
& \bloss{2.35}{-1.44} & \bloss{21.69}{-0.82} & \bloss{44.40}{-4.80} & \loss{16.89}{-14.88} \\
\cdashline{1-9}

gemma-4-E4B-it
& 0.67 & 16.05 & 49.35 & 40.01
& 1.39 & 17.86 & 53.72 & 31.48 \\
\quad +SFT w/o reasoning
& \gain{5.21}{4.54} & \gain{26.73}{10.68} & \gain{50.53}{1.18} & \loss{37.62}{-2.39}
& \gain{3.07}{1.68} & \gain{24.30}{6.44} & \loss{48.09}{-5.63} & \loss{25.21}{-6.27} \\
\quad +SFT w/ reasoning
& \gain{1.97}{1.30} & \gain{24.46}{8.41} & \loss{46.52}{-2.83} & \loss{23.95}{-16.06}
& \bgain{3.39}{2.00} & \gain{24.25}{6.39} & \bloss{50.30}{-3.42} & \loss{24.62}{-6.86} \\
\cdashline{1-9}

Qwen3-4B-Thinking
& 0.18 & 6.71 & 35.72 & 28.60
& 0.22 & 5.86 & 44.03 & 25.14 \\
\quad +SFT w/o reasoning
& \gain{3.05}{2.87} & \gain{18.63}{11.92} & \gain{42.63}{6.91} & \gain{30.63}{2.03}
& \gain{0.36}{0.14} & \gain{8.10}{2.24} & \gain{47.04}{3.01} & \gain{26.52}{1.38} \\
\quad +SFT w/ reasoning
& \bgain{4.95}{4.77} & \bgain{25.93}{19.22} & \bgain{50.91}{15.19} & \bgain{32.02}{3.42}
& \bgain{3.23}{3.01} & \bgain{24.22}{18.36} & \bgain{53.12}{9.09} & \gain{26.26}{1.12} \\
\cdashline{1-9}

Qwen3-8B
& 2.08 & 26.69 & 48.43 & 35.97
& 3.28 & 20.84 & 46.23 & 24.43 \\
\quad +SFT w/o reasoning
& \gain{3.92}{1.84} & \loss{23.87}{-2.82} & \loss{46.18}{-2.25} & \loss{31.38}{-4.59}
& \gain{4.02}{0.74} & \gain{23.55}{2.71} & \gain{52.47}{6.24} & \gain{28.69}{4.26} \\
\quad +SFT w/ reasoning
& \bgain{4.57}{2.49} & \bloss{25.76}{-0.93} & \bgain{48.98}{0.55} & \bloss{35.70}{-0.27}
& \bgain{4.13}{0.85} & \bgain{25.13}{4.29} & \bgain{53.35}{7.12} & \gain{26.44}{2.01} \\

\bottomrule
\end{tabular}%
}
\caption{SFT performance on \texttt{sjo} and \texttt{ctn}. Changes relative to the corresponding pretrained baseline are shown in parentheses. \textbf{Bold} indicates cases where SFT with reasoning traces outperforms SFT without reasoning traces. Overall, SFT with reasoning tends to outperform SFT without reasoning, although the effect is mixed.}
\label{tab:sft-results}
\end{table*}

As shown in \Cref{tab:sft-results}, \textbf{the effect of including reasoning traces is not consistent across metrics and models}, although the overall trend is that SFT with reasoning tends to outperform SFT without reasoning more often than the reverse.

The greatest improvements are observed for Qwen3-4B-Thinking-2507. \textit{SFT with reasoning} yields substantial gains over the unfine-tuned baseline on both \texttt{sjo} (+4.77 BLEU, +19.22 chrF, +15.19 SBERT, and +3.42 LLMaJ) and \texttt{ctn} (+3.01 BLEU, +18.36 chrF, +9.09 SBERT, and +1.12 LLMaJ). 

However, \textit{SFT without reasoning} also achieves strong gains in this case, indicating that the \textbf{improvements cannot be attributed solely to the inclusion of reasoning traces, but also arise from fine-tuning on the final translations.} Moreover, Qwen3-4B-Thinking-2507 has a relatively low baseline, so larger gains do not necessarily correspond to high final performance.

Compared with the ICL results, where reasoning traces provide large and consistent gains, the SFT results suggest that incorporating reasoning traces into training data is less beneficial than using them as in-context guidance.

Manual inspection of the generated responses shows that, after a few hundred initial training steps, the models can readily reproduce the format and style of the reasoning traces used for training. However, the actual reasoning content often still contains errors, which limits further improvement in the final translations (see \Cref{sec:output_examples} for an example).

\section{Reinforcement Fine-Tuning Experiment}
Although SFT does not yield consistent gains, the fine-tuned models learn to reliably produce step-by-step linguistic reasoning in the required format, providing a suitable starting point for RL. We therefore conduct a RFT experiment to test whether RFT can further improve models that have already been SFT-trained with reasoning traces.

\subsection{Setup}
We continue training from the previously SFT-trained LoRA adapters using Group Relative Policy Optimization (GRPO)~\citep{shao2024deepseekmathpushinglimitsmathematical}. For Qwen3-4B, we sample 8 completions per prompt with an effective batch size of 128. For Qwen3-8B, gemma-4-E2B-it, and gemma-4-E4B-it, we sample 4 completions per prompt with an effective batch size of 64 due to higher memory requirements.
More implementation details are provided in \Cref{sec:rft-setting}.

\subsection{Reward Functions}\label{sec:reward_function}
For reward functions, we combine MT metrics with rule-based format checks \citep{feng-etal-2025-mt-r1}. The translation reward is computed between the generated translation and the reference using sentence-level chrF, sentence-level BLEU, and SBERT, with weights 0.55, 0.15, and 0.25, respectively.\footnote{Sentence-level BLEU is assigned a smaller weight since it is less reliable than corpus-level BLEU \citep{chen-cherry-2014-systematic}.} An exact match receives a bonus of 0.05.

The format reward encourages the required output structure: a \texttt{<think>} block containing at least one \textit{Step}, followed by an \texttt{<answer>} block. 

We additionally use a process reward for the bracketed partial translations in the intermediate reasoning, based on recall-heavy matching between the lists of generated and gold partial translations using a combination of exact match, chrF, and SBERT. 

The top-level weights are 0.75 for the translation reward, 0.10 for the format reward, and 0.15 for the process reward. Further details are provided in \Cref{sec:reward_details}.

\subsection{Results and Analysis}

\begin{table*}[t]
\centering
\small
\setlength{\tabcolsep}{4pt}
\renewcommand{\arraystretch}{1.12}
\newcommand{\gain}[2]{\textbf{#1} {\scriptsize\textcolor{teal}{(+#2)}}}
\newcommand{\loss}[2]{#1 {\scriptsize\textcolor{gray}{(#2)}}}
\resizebox{\textwidth}{!}{%
\begin{tabular}{@{}lcccc:cccc@{}}
\toprule
\multirow{2}{*}{Model}
& \multicolumn{4}{c:}{Xibe (\texttt{sjo})}
& \multicolumn{4}{c}{Chintang (\texttt{ctn})} \\
\cmidrule(lr){2-5}\cmidrule(lr){6-9}
& BLEU & chrF & SBERT & LLMaJ & BLEU & chrF & SBERT & LLMaJ \\
\midrule

gemma-4-E2B-it (SFT-ed w/ reasoning)
& 1.24 & 20.71 & 37.44 & 25.64
& 2.35 & 21.69 & 44.40 & 16.89 \\
\quad +RFT
& \gain{1.84}{0.60} & \gain{21.45}{0.74} & \gain{40.51}{3.07} & \gain{25.82}{0.18}
& \loss{2.09}{-0.26} & \gain{21.71}{0.02} & \gain{45.24}{0.84} & \gain{18.98}{2.09} \\
\cdashline{1-9}

gemma-4-E4B-it (SFT-ed w/ reasoning)
& 1.97 & 24.46 & 46.52 & 23.95
& 3.39 & 24.25 & 50.30 & 24.62 \\
\quad +RFT
& \gain{3.41}{1.44} & \loss{23.35}{-1.11} & \loss{45.44}{-1.08} & \gain{28.23}{4.28}
& \gain{3.41}{0.02} & \loss{23.01}{-1.24} & \loss{48.29}{-2.01} & \gain{24.69}{0.07} \\
\cdashline{1-9}

Qwen3-4B-Thinking (SFT-ed w/ reasoning)
& 4.95 & 25.93 & 50.91 & 32.02
& 3.23 & 24.22 & 53.12 & 26.26 \\
\quad +RFT
& \loss{4.11}{-0.84} & \loss{24.97}{-0.96} & \loss{48.81}{-2.10} & \loss{30.22}{-1.80}
& \gain{3.48}{0.25} & \loss{24.15}{-0.07} & \loss{52.23}{-0.89} & \loss{24.96}{-1.30} \\
\cdashline{1-9}

Qwen3-8B (SFT-ed w/ reasoning)
& 4.57 & 25.76 & 48.98 & 35.70
& 4.13 & 25.13 & 53.35 & 26.44 \\
\quad +RFT
& \loss{3.73}{-0.84} & \gain{26.21}{0.45} & \gain{49.53}{0.55} & \loss{32.40}{-3.30}
& \loss{4.00}{-0.13} & \gain{25.51}{0.38} & \gain{53.56}{0.21} & \loss{25.55}{-0.89} \\

\bottomrule
\end{tabular}%
}
\caption{Performance comparison between models SFT-trained with reasoning traces and further RFT-ed models on \texttt{sjo} and \texttt{ctn}. Gains from RFT are shown in parentheses. \textbf{Bold} indicates scores higher than the baseline before SFT. RFT yields no clear gains over SFT.}
\label{tab:rft-results}
\end{table*}

As shown in \Cref{tab:rft-results}, \textbf{RFT leads to only small changes in performance}, with both gains and degradations remaining very limited across metrics. The effects are mixed but similarly small across metrics on both \texttt{sjo} and \texttt{ctn}. These results suggest that, under the current RFT setup, reinforcement fine-tuning does not yield substantial improvements beyond the SFT models trained with reasoning traces.

Manual inspection of the generated responses reveals a pattern similar to that observed in SFT: the models learn to produce step-by-step linguistic reasoning in the expected way, but the actual reasoning content often remains incorrect (see \Cref{sec:output_examples} for an example). The models frequently analyze sentence structure incorrectly, assign incorrect dependency relations between words, and fail to select the appropriate senses for polysemous words. This suggests that lacking knowledge to correctly analyze low-resource languages may be the main bottleneck.

Overall, the performance of RFT on top of SFT still lags far behind the ICL setting. The higher performance in ICL can be attributed to the fact that the reasoning traces used in ICL are generated from gold-standard annotations and therefore provide reliable guidance for analyzing the linguistic structure of each sentence. In contrast, models trained with SFT and RFT must generate the linguistic analysis themselves, and they still often fail to do so correctly. Such incorrect analyses propagate to the final translations and limit final translation quality.

Another limiting factor is that our RL setup may not yet provide sufficient exploration. Due to computational constraints, we use a relatively small number of sampled generations per prompt, which limits the model's exploration space. This limitation may be important for linguistic reasoning, where each sentence can potentially be analyzed in many different ways. As a result, the search space may be too large for the current RL setup to reliably discover and reinforce correct reasoning trajectories.

\section{Conclusion}

In this work, we develop a pipeline for automatically generating linguistic reasoning traces and evaluate their effectiveness for low-resource MT in three settings: ICL, SFT, and RFT, each comparing against a corresponding baseline without the reasoning traces.
Our results show that these traces are most effective when used as in-context guidance: they provide reliable sentence-specific analyses and substantially improve translation performance. 
In contrast, using the same traces as training data yields smaller and less consistent gains, as models can learn to reproduce the trace format but still often generate error-prone reasoning content, limiting its effectiveness in improving final translation quality.
Further RFT does not bring meaningful improvements over SFT. 
Overall, our findings suggest that LLMs can leverage grammatical information for low-resource MT when provided with reliable linguistic analyses, but learning to generate such analyses remains a key bottleneck.

\section*{Limitations}

Our ICL approach relies on gold-standard UD annotations or a sufficiently accurate UD parser, which are unavailable for many low-resource languages. This limitation, however, motivates the continued expansion of UD treebanks and the development of reliable UD parsers for more low-resource languages.

Our RFT experiment is limited by computational constraints, and we use a relatively small number of sampled generations per prompt and a limited batch size, which restricts the exploration space available to the model during RL training.
Therefore, the limited gains observed in our RFT experiments may be interpreted as  our current RL setup being not sufficient for the models to reliably explore and discover correct linguistic reasoning trajectories. We have explored only a limited set of reward weight configurations, and a more systematic ablation of the reward components and weights would be valuable in future work.

A second limitation is that our current reward function is based primarily on MT metrics and does not directly reward syntactic analysis. 
Although we incorporate an intermediate process reward, this verification still only checks surface-level phrasal translations rather than the syntactic analysis itself. As a result, the reward signal may be too weak to teach the model accurate linguistic reasoning. 

In future work, we could extract not only intermediate phrasal translations from the model's reasoning, but also its predicted dependency analyses, and verify them against the gold UD tree structures. This could provide a stronger reward signal for learning syntactic analysis. Once the syntactic analysis becomes more accurate, models will be in a much better position to exploit grammatical information for downstream translation, which could lead to improvements as observed in the ICL experiment.

\section*{Ethical Considerations}

\paragraph{Use of AI Assistants.}
The authors used ChatGPT for grammar correction, clarity improvement, and coherence polishing, and OpenAI Codex for assistance with code implementations. The authors retain full responsibility for all technical contributions, experimental design decisions, analyses, and the final content of the paper.\footnote{ChatGPT: \url{https://chatgpt.com/}; OpenAI Codex: \url{https://chatgpt.com/codex/}.}

\section*{Acknowledgments}
The authors wish to acknowledge CSC – IT Center for Science, Finland, for computational resources.
We thank Siyao Peng for his contributions and invaluable feedback. We also thank Fresco Sam-Sin of the Manchu Foundation for generously granting us permission to use the digitized Manchu materials available on his website.
Yihong Liu and Hinrich Sch\"utze were supported by the Munich Center for Machine Learning (MCML) and German Research Foundation (DFG, grant SCHU 2246/14-1).
Sampo Pyysalo received funding from the Digital Europe Programme under grant agreement No 101195233 (OpenEuroLLM).
Shaoxiong Ji gratefully acknowledges the support of Foundation PS through the PS Fellowship.

\bibliography{custom}

\appendix

\section{Implementation Details}\label{sec:implementation_details}
Our experiments used approximately 2,000 GPU-hours on AMD MI250X GPUs.
\subsection{In-Context Learning Experiment}
\label{sec:icl-setting}
For decoding hyperparameters, we follow the recommendations in the respective model cards, using a temperature of 1.0, nucleus sampling with $p=0.95$, and top-$k$ sampling with $k=64$ for Gemma 4 models, and a temperature of 0.6, nucleus sampling with $p=0.95$, and top-$k$ sampling with $k=20$ for Qwen 3 models.

For the Qwen3 models, we set \texttt{enable\_thinking=True}, which applies the models' native chat template for thinking mode. For the Gemma 4 models, our pilot study shows that enabling a thinking template causes the models to generate additional reasoning outside our designated \texttt{<think>} block, which lowers performance. We therefore set \texttt{enable\_thinking=False} for Gemma 4 models.

\subsection{Supervised Fine-Tuning Experiment}
\label{sec:sft-setting}
For SFT, we use low-rank adaptation (LoRA) parameter-efficient fine-tuning. Training is performed with a batch size of 8, bfloat16 precision, a learning rate of $1\times10^{-5}$, weight decay of 0.01, and a maximum of 2,000 optimization steps. The best checkpoint is selected based on evaluation loss on the held-out validation set. LoRA is applied with rank $r=16$, scaling factor $\alpha=8$, and dropout 0.05. The models are trained with completion-only loss, such that the loss is computed only over the target answer tokens.

\subsection{Reinforcement Fine-Tuning Experiment}
\label{sec:rft-setting}
For RFT, we use \texttt{trl} version 1.4.0 and \texttt{vLLM} version 0.20.1 in colocated mode for faster generation.
For decoding hyperparameters, we follow the recommendations in the respective model cards, using a temperature of 1.0, nucleus sampling with $p=0.95$, and top-$k$ sampling with $k=64$ for Gemma 4 models, and a temperature of 0.6, nucleus sampling with $p=0.95$, and top-$k$ sampling with $k=20$ for Qwen 3 models.

For Qwen3-4B, the RFT runs use 8 sampled completions per prompt, a distributed batch size of 16, and an effective optimization batch size of 128 after gradient accumulation. Due to the higher memory demands, the RFT runs of the larger Qwen3-8B model use 4 sampled completions per prompt, a distributed batch size of 8, and an effective optimization batch size of 64. 

All runs use bfloat16 precision, learning rate of $1\times10^{-6}$, and LoRA with rank $r=16$, scaling factor $\alpha=8$, and dropout 0.05. We train the models for 600 steps, save checkpoints every 100 steps, and select the best checkpoint using the validation set.

\section{Reward Function Details}\label{sec:reward_details}
The reward function used in the RFT experiment is a weighted sum of three rewards, with top-level weights of 0.75 for the translation reward, 0.10 for the format reward, and 0.15 for the process reward. The design makes final translation quality the dominant optimization target while still explicitly encouraging structural compliance and faithful intermediate reasoning.

\subsection{Final-translation reward}
The translation reward is computed from the generated final translation in the \texttt{<answer>} block and combines sentence-level chrF, sentence-level BLEU, and SBERT similarity with weights 0.55, 0.15, and 0.25, respectively. Sentence-level BLEU is assigned with a smaller weight, since it is less reliable than corpus-level BLEU. An exact-match receives a bonus of 0.05 and empty answers receive a penalty of 0.25. 

\subsection{Format reward}
The format reward encourages the required tagged output structure and assigns bonuses of 0.10 for the presence of a \texttt{<think>} block, 0.10 for the presence of an \texttt{<answer>} block, 0.05 for correct \texttt{<think>}-before-\texttt{<answer>} ordering, 0.10 for a non-empty answer, 0.10 for the presence of at least one explicit step marker, 0.03 if reasoning starts at Step 1, and 0.02 if step numbering is monotonic; penalties are applied for missing \texttt{<think>} (0.10), missing \texttt{<answer>} (0.20), empty tagged content (0.20), malformed step structure (0.10), wrong tag order (0.10), and trailing text after the final \texttt{</answer>} tag (0.05). 

\subsection{Partial-translation process reward}
We use a process reward defined over the intermediate reasoning trace in the \texttt{<think>} block. Partial translations are extracted from this block, and the resulting list of generated partial translations is compared against the list of gold partial translations. We compute both recall-oriented matching, where each gold phrase is matched to its best-matching generated phrase, and precision-oriented matching, where each generated phrase is matched to its best-matching gold phrase.

For short phrases of up to two tokens, phrase similarity is computed as a weighted combination of exact match and chrF, with weights 0.65 and 0.35, respectively. For longer phrases, phrase similarity combines exact match, chrF, and SBERT similarity, with weights 0.15, 0.70, and 0.15. These phrase-level similarities are then aggregated into the process reward.

We use a recall-heavy soft-matching objective, with weights of 0.75 for recall and 0.20 for precision, together with a non-empty prediction bonus of 0.05.

\section{Prompt Templates}\label{sec:prompt_templates}

\subsection{LLM-as-a-Judge Prompt}\label{sec:prompt_templates_llm_judge}

\begin{tcolorbox}[title=LLM-as-a-Judge prompt, breakable]
Score the following translation from \texttt{\{source\_lang\}} to English on a scale from 0 to 100 based on adequacy, where a score of 0 means that the translation conveys little or none of the reference meaning; 33 indicates a flawed translation with serious mistranslations, omissions, or additions; 66 indicates a translation that conveys the main meaning of the reference with only minor meaning errors or ambiguities; and 100 represents a translation that fully preserves the meaning of the reference. Answer with only a whole number representing the score, and nothing else.

\medskip
\noindent Translation: \texttt{\{translation\}}

\noindent Reference: \texttt{\{gold\_translation\}}
\end{tcolorbox}

\subsection{Prompts for LLM to Fill in Placeholders}\label{sec:prompt_templates_llm_fill_in_placeholder}

\begin{tcolorbox}[title=Prompt for LLM to Fill in Placeholders (Only Phrasal Translation), breakable]
You are given dictionary entries for each individual word in a \texttt{\{source\_lang\}} sentence, and a step-by-step reasoning process that explains how to combine the meanings of these individual words to phrases, and finally arrive at the meaning of the whole sentence.

\medskip
\noindent Your task is to complete the reasoning steps below by filling in every placeholder enclosed in square brackets, as in [Phrasal Translation].

\noindent In [Phrasal Translation], you should fill in the English translation of the complete phrase or clause formed at that step.

\medskip
\noindent Instructions:

\begin{enumerate}
    \item The dictionary entry may contain multiple possible meanings for a word; the specific lexical meaning in the context and its morphological features are already explained in the reasoning steps.
    \item Based on the explanations of each word, and the final English translation of the whole sentence provided at the end of the reasoning steps, your task is to fill in each [Phrasal Translation] by combining the meanings of the words according to the syntactic relations explained in the reasoning steps.
    \item Ensure consistency between the phrasal translation and the final English translation of the whole sentence.
    \item Preserve all original formatting, and keep all square brackets exactly as they appear.
    \item Do not add, remove, or modify any text outside of filling in the placeholders of [Phrasal Translation].
    \item Return only the completed reasoning steps. Do not include explanations, comments, or additional text.
\end{enumerate}

\noindent Dictionary entries:

\noindent \texttt{\{dictionary\_entries\}}

\medskip
\noindent Reasoning steps with placeholders:

\noindent \texttt{\{reasoning\_steps\_with\_placeholders\}}
\end{tcolorbox}

\begin{tcolorbox}[title=Prompt for LLM to Fill in Placeholders (Lexical Meaning and Phrasal Translation), breakable]
You are given dictionary entries for each individual word in a \texttt{\{source\_lang\}} sentence, and a step-by-step reasoning process that explains how to combine the meanings of these individual words to arrive at the meaning of the whole sentence.

\medskip
\noindent Your task is to complete the reasoning steps below by filling in every placeholder enclosed in square brackets (e.g., [Lexical Meaning], [Phrasal Translation]).

\noindent In [Lexical Meaning], you should fill in the most contextually appropriate English meaning of the individual \texttt{\{source\_lang\}} word. The square brackets should be kept.

\noindent In [Phrasal Translation], you should fill in the English translation of the complete phrase or clause formed at that step. The square brackets should be kept.

\medskip
\noindent Instructions:

\begin{enumerate}
    \item Use only the provided dictionary entries as your source of lexical meanings.
    \item If a dictionary entry contains multiple possible meanings, select the one that best fits the final English sentence provided at the end of the reasoning steps.
    \item Ensure consistency between the chosen lexical meanings and the final sentence translation.
    \item Preserve all original formatting, and keep the square brackets after filling in the placeholders.
    \item Do not add, remove, or modify any text outside of filling in the placeholders.
    \item Return only the completed reasoning steps. Do not include explanations, comments, or additional text.
\end{enumerate}

\noindent Dictionary entries:

\noindent \texttt{\{dictionary\_entries\}}

\medskip
\noindent Reasoning steps with placeholders:

\noindent \texttt{\{reasoning\_steps\_with\_placeholders\}}
\end{tcolorbox}

\subsection{In-Context MT Prompts}\label{sec:prompt_templates_in_context_mt}

\begin{tcolorbox}[title=Baseline In-Context MT Prompt, breakable]
Please help me translate the following sentence from \texttt{\{source\_lang\}} to English:

\noindent \texttt{\{source\_sentence\}}

\medskip
\noindent For the translation task, you are given the dictionary entries for each individual word of the \texttt{\{source\_lang\}} sentence.
Some words may be polysemous and there might be multiple possible English translations. In such cases, please choose the most appropriate one.

\medskip
\noindent Here are the dictionary entries for each individual word in the source sentence:

\noindent \texttt{\{dictionary\_entries\}}

\medskip
\noindent You are also given the grammar rules that are directly relevant to this sentence:

\noindent \texttt{\{relevant\_grammar\_rules\}}

\medskip
\noindent Using all the information provided above, you should proceed step-by-step: first determine the meaning and part-of-speech of each word; then identify the syntactic relationships among the words; then based on the syntactic relationships, combine the meanings of individual words to get phrase meanings; and continue this process until you eventually derive the meaning of the whole sentence.

\medskip
\noindent Your response must contain exactly two sections:

\noindent 1. Step-by-step reasoning inside \texttt{<think> ... </think>}

\noindent 2. The final English translation inside \texttt{<answer> ... </answer>}

\noindent Do not add any extra text outside these tags.

\medskip
\noindent Remember your source sentence is: \texttt{\{source\_sentence\}}
\end{tcolorbox}

\begin{tcolorbox}[title=\textit{+Reasoning} In-Context MT prompt, breakable]
Please help me translate the following sentence from \texttt{\{source\_lang\}} to English:

\noindent \texttt{\{source\_sentence\}}

\medskip
\noindent Dictionary entries:

\noindent \texttt{\{dictionary\_entries\}}

\medskip
\noindent You are also given a linguistic reasoning guide for this sentence.
Use it as in-context guidance for translating the sentence.
Some placeholders for Lexical Meanings, Phrasal Translations, and the Final Translation are intentionally left unfilled.

\medskip
\noindent Linguistic reasoning guide:

\noindent \texttt{\{reasoning\_trace\_with\_placeholders\}}

\medskip
\noindent Your task is to use the dictionary entries and the linguistic reasoning guide above to derive the final English translation.
Proceed step by step through the linguistic reasoning guide and resolve the placeholders in ascending order.

\medskip
\noindent For each lexical-meaning placeholder, such as [Lexical Meaning 1]:
\begin{enumerate}
    \item Use the provided dictionary entries as the source of possible word meanings.
    \item Choose the meaning that best fits the local context.
    \item Use the word explanation immediately before the placeholder, including part of speech, lemma, morphology, case, tense, aspect, number, person, polarity, or other grammatical features.
    \item If the dictionary gives multiple meanings, prefer the one that is compatible with the morphological and syntactic explanation in the guide.
    \item Sometimes there is no dictionary entry for a word. In that case, try to guess the meaning based on the word form, the context, and the reasoning guide. It could be a proper noun, a loanword, a compound, or a typo.
\end{enumerate}

\noindent For each phrasal-translation placeholder, such as [Phrasal Translation 1]:
\begin{enumerate}
    \item Combine meanings that have already been resolved in earlier placeholders.
    \item Use the syntactic relationship described in the guide to decide how the dependent combines with the head.
    \item Use word order, case marking, adpositions, auxiliaries, modifiers, subjects, objects, clauses, and any provided grammar notes.
    \item Translate the whole subtree named in that line, not just the head word.
    \item Make the phrase meaning consistent with the meanings chosen for its component words.
\end{enumerate}

\noindent Continue this bottom-up process until you reach [Final Translation].

\medskip
\noindent Important output requirements:

\noindent Output first the completed linguistic reasoning guide with all placeholders resolved inside \texttt{<completed\_guide> ... </completed\_guide>} tags.

\noindent Then output only the final English translation inside \texttt{<answer> ... </answer>} tags.

\noindent Do not add any text before \texttt{<completed\_guide>}, between \texttt{</completed\_guide>} and \texttt{<answer>}, or after \texttt{</answer>}.
\end{tcolorbox}

\section{Intermediate Reasoning Correctness}
\label{sec:intermediate_correctness}

To examine whether erroneous intermediate reasoning corresponds to lower final translation quality, we conduct an additional group-based analysis of the ICL outputs from the best-performing gemma-4-31B-it model. For each sentence, we evaluate whether every generated intermediate phrasal translation matches its gold-standard counterpart, using exact matching supplemented by an LLM-as-a-judge assessment. We then divide the outputs into two groups: (i) those for which all intermediate phrasal translations are correct, and (ii) those containing at least one intermediate translation error. For each group, we report the number of sentences and the average LLMaJ score of the final translation.

\begin{table}[h]
\centering
\small
\setlength{\tabcolsep}{4pt}
\renewcommand{\arraystretch}{1.08}
\begin{tabular}{@{}lrrrrr@{}}
\toprule
Group & \# Sent. & BLEU & chrF & SBERT & LLMaJ \\
\midrule
\multicolumn{6}{@{}l}{\textbf{Xibe}} \\
Fully correct & 69  & 18.42 & 44.35 & 70.65 & 74.39 \\
Has error   & 82  & 6.17  & 33.23 & 58.93 & 53.95 \\
All & 151 & 10.81 & 37.36 & 64.28 & 63.29 \\
\midrule
\multicolumn{6}{@{}l}{\textbf{Chintang}} \\
Fully correct & 219 & 12.61 & 37.89 & 77.66 & 72.81 \\
Has error   & 125 & 9.74  & 34.46 & 66.11 & 51.70 \\
All & 344 & 12.59 & 36.53 & 73.46 & 65.14 \\
\bottomrule
\end{tabular}
\caption{Final translation quality by intermediate phrasal-translation correctness in the ICL +reasoning setting of gemma-4-31B-it. Sentences are grouped by whether all intermediate phrasal translations are correct or at least one is erroneous. ``\# Sent.'' denotes the number of sentences in each group.}
\label{tab:intermediate_correctness}
\end{table}

As shown in \Cref{tab:intermediate_correctness}, outputs with fully correct intermediate phrasal translations achieve substantially higher final translation quality across all four metrics. Compared with outputs containing at least one erroneous intermediate phrasal translation, the all-correct group achieves higher scores by +12.25 BLEU, +11.12 chrF, +11.72 SBERT, and +20.44 LLMaJ for Xibe, and by +2.87 BLEU, +3.43 chrF, +11.55 SBERT, and +21.11 LLMaJ for Chintang. It shows that correct intermediate phrasal-translation steps consistently lead to better final translations, and conversely, errors accumulated in the intermediate steps will degrade the final translation.

\section{Dictionary-Only ICL Baseline}
\label{sec:dictionary_only_baseline}

To see how ICL performs without any grammatical information, we conduct an additional dictionary-only ICL experiment in which the model receives only the relevant dictionary entries but no grammar rules or reasoning guide. We compare this dictionary-only setting against the baseline with the triggered grammar rules as a flat list, and the +reasoning setting with the same linguistic rules organized into the step-by-step reasoning structure. 

\begin{table*}[t]
\centering
\small
\setlength{\tabcolsep}{4pt}
\renewcommand{\arraystretch}{1.08}
\newcommand{\gain}[2]{#1 {\scriptsize\textcolor{teal}{(+#2)}}}
\newcommand{\loss}[2]{#1 {\scriptsize\textcolor{gray}{(#2)}}}
\begin{tabular}{@{}lllrrrr@{}}
\toprule
Model & Language & Setting & BLEU & chrF & SBERT & LLMaJ \\
\midrule

\multirow{6}{*}{gemma-4-E2B-it}
& \multirow{3}{*}{Xibe}
& dictionary only
& \textbf{1.64} & \textbf{27.18} & 49.51 & 34.48 \\
& & w/ grammar rules
& \loss{0.76}{-0.88}
& \loss{22.15}{-5.03}
& \loss{47.57}{-1.94}
& \gain{\textbf{42.87}}{8.39} \\
& & +reasoning
& \loss{0.58}{-1.06}
& \loss{17.49}{-9.69}
& \gain{\textbf{51.30}}{1.79}
& \gain{39.48}{5.00} \\
\cmidrule(lr){2-7}
& \multirow{3}{*}{Chintang}
& dictionary only
& 3.20 & 21.46 & 45.05 & 23.86 \\
& & w/ grammar rules
& \gain{\textbf{3.79}}{0.59}
& \gain{22.51}{1.05}
& \gain{49.20}{4.15}
& \gain{31.77}{7.91} \\
& & +reasoning
& \gain{3.37}{0.17}
& \gain{\textbf{24.08}}{2.62}
& \gain{\textbf{51.26}}{6.21}
& \gain{\textbf{41.54}}{17.68} \\
\midrule

\multirow{6}{*}{gemma-4-E4B-it}
& \multirow{3}{*}{Xibe}
& dictionary only
& \textbf{2.99} & \textbf{27.38} & 50.32 & 36.81 \\
& & w/ grammar rules
& \loss{0.67}{-2.32}
& \loss{16.05}{-11.33}
& \loss{49.35}{-0.97}
& \gain{40.01}{3.20} \\
& & +reasoning
& \loss{1.88}{-1.11}
& \loss{22.70}{-4.68}
& \gain{\textbf{54.49}}{4.17}
& \gain{\textbf{45.38}}{8.57} \\
\cmidrule(lr){2-7}
& \multirow{3}{*}{Chintang}
& dictionary only
& 3.97 & 23.07 & 51.68 & 25.84 \\
& & w/ grammar rules
& \loss{1.39}{-2.58}
& \loss{17.86}{-5.21}
& \gain{53.72}{2.04}
& \gain{31.48}{5.64} \\
& & +reasoning
& \gain{\textbf{6.96}}{2.99}
& \gain{\textbf{29.75}}{6.68}
& \gain{\textbf{65.04}}{13.36}
& \gain{\textbf{49.99}}{24.15} \\
\midrule

\multirow{6}{*}{gemma-4-31B-it}
& \multirow{3}{*}{Xibe}
& dictionary only
& 8.26 & 34.57 & 60.10 & 55.68 \\
& & w/ grammar rules
& \gain{9.84}{1.58}
& \gain{35.85}{1.28}
& \gain{62.31}{2.21}
& \gain{59.69}{4.01} \\
& & +reasoning
& \gain{\textbf{10.81}}{2.55}
& \gain{\textbf{37.36}}{2.79}
& \gain{\textbf{64.28}}{4.18}
& \gain{\textbf{63.29}}{7.61} \\
\cmidrule(lr){2-7}
& \multirow{3}{*}{Chintang}
& dictionary only
& 7.74 & 28.79 & 61.29 & 37.53 \\
& & w/ grammar rules
& \gain{9.96}{2.22}
& \gain{31.61}{2.82}
& \gain{63.13}{1.84}
& \gain{43.24}{5.71} \\
& & +reasoning
& \gain{\textbf{12.59}}{4.85}
& \gain{\textbf{36.53}}{7.74}
& \gain{\textbf{73.46}}{12.17}
& \gain{\textbf{65.14}}{27.61} \\
\midrule

\multirow{6}{*}{Qwen3-4B-Thinking}
& \multirow{3}{*}{Xibe}
& dictionary only
& 0.36 & 10.35 & 39.56 & \textbf{44.49} \\
& & w/ grammar rules
& \loss{0.18}{-0.18}
& \loss{6.71}{-3.64}
& \loss{35.72}{-3.84}
& \loss{28.60}{-15.89} \\
& & +reasoning
& \gain{\textbf{1.21}}{0.85}
& \gain{\textbf{17.54}}{7.19}
& \gain{\textbf{53.21}}{13.65}
& \loss{41.82}{-2.67} \\
\cmidrule(lr){2-7}
& \multirow{3}{*}{Chintang}
& dictionary only
& 0.49 & 9.63 & 46.59 & 34.83 \\
& & w/ grammar rules
& \loss{0.22}{-0.27}
& \loss{5.86}{-3.77}
& \loss{44.03}{-2.56}
& \loss{25.14}{-9.69} \\
& & +reasoning
& \gain{\textbf{1.11}}{0.62}
& \gain{\textbf{13.12}}{3.49}
& \gain{\textbf{63.77}}{17.18}
& \gain{\textbf{48.56}}{13.73} \\
\midrule

\multirow{6}{*}{Qwen3-8B}
& \multirow{3}{*}{Xibe}
& dictionary only
& 2.15 & 26.77 & 46.62 & 28.66 \\
& & w/ grammar rules
& \loss{2.08}{-0.07}
& \loss{26.69}{-0.08}
& \gain{48.43}{1.81}
& \gain{35.97}{7.31} \\
& & +reasoning
& \gain{\textbf{2.74}}{0.59}
& \gain{\textbf{27.79}}{1.02}
& \gain{\textbf{55.57}}{8.95}
& \gain{\textbf{43.23}}{14.57} \\
\cmidrule(lr){2-7}
& \multirow{3}{*}{Chintang}
& dictionary only
& 1.72 & 17.45 & 38.58 & 45.26 \\
& & w/ grammar rules
& \gain{3.28}{1.56}
& \gain{20.84}{3.39}
& \gain{46.23}{7.65}
& \loss{24.43}{-20.83} \\
& & +reasoning
& \gain{\textbf{7.02}}{5.30}
& \gain{\textbf{31.19}}{13.74}
& \gain{\textbf{64.80}}{26.22}
& \gain{\textbf{45.55}}{0.29} \\
\midrule

\multirow{6}{*}{Qwen3-14B}
& \multirow{3}{*}{Xibe}
& dictionary only
& 3.41 & 28.33 & 51.68 & 38.17 \\
& & w/ grammar rules
& \loss{3.11}{-0.30}
& \gain{28.99}{0.66}
& \gain{52.79}{1.11}
& \gain{43.56}{5.39} \\
& & +reasoning
& \gain{\textbf{4.70}}{1.29}
& \gain{\textbf{30.83}}{2.50}
& \gain{\textbf{57.09}}{5.41}
& \gain{\textbf{48.74}}{10.57} \\
\cmidrule(lr){2-7}
& \multirow{3}{*}{Chintang}
& dictionary only
& 3.71 & 25.26 & 53.78 & 30.34 \\
& & w/ grammar rules
& \gain{5.25}{1.54}
& \gain{26.50}{1.24}
& \gain{56.10}{2.32}
& \gain{32.91}{2.57} \\
& & +reasoning
& \gain{\textbf{10.33}}{6.62}
& \gain{\textbf{34.07}}{8.81}
& \gain{\textbf{67.07}}{13.29}
& \gain{\textbf{51.98}}{21.64} \\
\bottomrule
\end{tabular}

\caption{ICL performance comparison between dictionary entries only, dictionary entries plus triggered grammar rules, and UD-derived reasoning guides (\textit{+reasoning}). Changes relative to the dictionary-only baseline are shown in parentheses. \textbf{Bold} indicates the highest score among the three settings for each model, language, and metric.}
\label{tab:dictionary_only_icl}
\end{table*}

As shown by the results in \Cref{tab:dictionary_only_icl}, simply adding triggered grammar rules to the dictionary entries does not consistently improve performance, particularly for smaller models, and can sometimes underperform the dictionary-only setting. In contrast, organizing the grammatical information into a sentence-specific, step-by-step reasoning process generally yields stronger performance, with +reasoning achieving the best results across most models, languages, and metrics. This suggests that the way grammatical information is structured and presented is important for effectively leveraging it in ICL.

\clearpage
\onecolumn
\section{Example of Erroneous Reasoning}\label{sec:output_examples}

\begin{table}[H]
\centering
\scriptsize
\setlength{\tabcolsep}{6pt}
\renewcommand{\arraystretch}{1.8}

\begin{tabularx}{\textwidth}{@{}XX@{}}
\toprule
\multicolumn{2}{c}{\textbf{Source Sentence:} \textit{xan be gidame honggon hvlhambi}
\quad \textbf{Sentence ID:} \texttt{textbook5-sjo\_133}} \\
\midrule
\textbf{Generated Reasoning Trace} & \textbf{Gold Reasoning Trace} \\
\midrule

\textbf{Step 1:}
Step 1:
The word 'hvlhambi' is a verb. Its lemma form is 'hvlhambi', which means [to rob, to steal]. 'hvlhambi' is present tense finite form.
According to the grammar of Xibe:
The form in -mbi is used to denote the present and future tenses. It is also used to indicate usual and frequentative or abstract actions...

 - So \err{'hvlhambi' means: [is robbing]}{wrong lexical sense for polysemous word}
 
The word 'honggon' is a noun. 

 - So 'honggon' means: [small bell]
&
\textbf{Step 1:}
The word 'xan' is a noun. 

 - So 'xan' means: [ear]
 
The word 'be' is a postposition. 'be' is accusative case marker.
According to the grammar of Xibe:
...The main syntactic function of the accusative is to denote a direct complement of the verb. According to this main role in a sentence, the accusative is used to indicate the direct object of a transitive verb...
According to the grammar of Xibe:
...case markers may attach syntactically to nouns, participles, and clauses.
As the postposition 'be' follows the noun 'xan', the syntactic relationship here is: 'be' is the case marking of 'xan'.

 - So \textcolor{blue}{'xan be' means: [ear (accusative)]}
\\[4pt]

\textbf{Step 2:}
The word 'gidame' is a verb. Its lemma form is 'gidambi', \err{which means [to press, to crush, to roll flat]}{wrong lexical sense for polysemous word}. 'gidame' is imperfective aspect converb form.
According to the grammar of Xibe:
The imperfect converb is formed with the suffix -me. Its main meaning is simultaneity between a dependent action and the principal action, and this relative temporal relation may occur in the past, present, or future...

 - So 'gidame' means: [pressing]
 
According to the grammar of Xibe:
...The converb is the predicate of the adverbial clause, and it is dependent on the main predicate, their relation is advcl...
As the verb 'gidame' precedes the verb 'hvlhambi', the syntactic relationship here is: \err{'gidame' is the adverbial clause modifier of 'hvlhambi'.}{wrong composition of phrase}

 - So 'gidame honggon hvlhambi' means: [pressing small bell is robbing]
&
\textbf{Step 2:}
The word 'gidame' is a verb. Its lemma form is 'gidambi', \textcolor{blue}{which means [to close, to shut, to turn off]}. 'gidame' is imperfective aspect converb form.
According to the grammar of Xibe:
The imperfect converb is formed with the suffix -me. Its main meaning is simultaneity between a dependent action and the principal action, and this relative temporal relation may occur in the past, present, or future...

 - So 'gidame' means: [closing/shutting]
 
According to the grammar of Xibe:
The main syntactic function of the accusative is to denote a direct complement of the verb. The subject appears in the nominative, and the object is marked by the accusative; however, when an inanimate direct object is close to the verb and is clearly the object of the action, there is no necessity in using the accusative.
As the noun 'xan' precedes the verb 'gidame', the syntactic relationship here is: 'xan be' is the object of 'gidame'.

 - So 'xan be gidame' means: [closing the ears]
\\[4pt]

\textbf{Step 3:}
The word 'xan' is a noun. 
 - So 'xan' means: [ear]
The word 'be' is a postposition. 'be' is accustative case marker.
According to the grammar of Xibe:
...The main syntactic function of the accusative is to denote a direct complement of the verb...
According to the grammar of Xibe:
...case markers may attach syntactically to nouns, participles, and clauses.
As the postposition 'be' follows the noun 'xan', the syntactic relationship here is: 'be' is the case marking of 'xan'.

 - So \err{'xan be' means: [at the ear]}{wrong interpretation of postposition}

\textbf{Step 4:}
The word 'honggon' is a noun. 

 - So 'honggon' means: [small bell]

As the noun 'honggon' precedes the verb 'hvlhambi', the syntactic relationship here is: \err{'honggon' is the compound of 'hvlhambi'}{wrong syntactic relation}.

 - So 'gidame honggon hvlhambi' means: [pressing small bell is robbing]

The word 'xan' is a noun. 

 - So 'xan' means: [ear]
 
The word 'be' is a postposition. 'be' is accustative case marker.
According to the grammar of Xibe:
...the accusative is used to indicate the direct object of a transitive verb...
As the postposition 'be' follows the noun 'xan', the syntactic relationship here is: 'be' is the case marking of 'xan'.
The word 'xan' is a noun. 

 - So 'xan be' means: [at the ear]
 
According to the grammar of Xibe:
...when an inanimate direct object is close to the verb and is clearly the object of the action, there is no necessity in using the accusative.
As the noun 'xan' precedes the verb 'hvlhambi', the syntactic relationship here is: 'xan be' is the object of 'hvlhambi'.
Putting all these pieces together, the whole sentence 'xan be gidame honggon hvlhambi' translates to: 'He is robbing small bells at the ear.'
 
&
\textbf{Step 3:}
The word 'hvlhambi' is a verb. 'hvlhambi' is present tense finite form. Its lemma form is 'hvlhambi', which means [to rob, to steal].
According to the grammar of Xibe:
The form in -mbi is used to denote the present and future tenses...

 - So \textcolor{blue}{'hvlhambi' means: [steals]}
 
According to the grammar of Xibe:
...The converb is the predicate of the adverbial clause...
As the verb 'gidame' precedes the verb 'hvlhambi', the syntactic relationship here is: \textcolor{blue}{'xan be gidame' is the adverbial clause modifier of 'hvlhambi'}.
The word 'honggon' is a noun. 

 - So 'honggon' means: [small bell]
 
According to the grammar of Xibe:
...when an inanimate direct object is close to the verb and is clearly the object of the action, there is no necessity in using the accusative.
As the noun 'honggon' precedes the verb 'hvlhambi', the syntactic relationship here is: \textcolor{blue}{'honggon' is the object of 'hvlhambi'}.
Putting all these pieces together, the whole sentence 'xan be gidame honggon hvlhambi' translates to: 'Cover ears while stealing a bell'
\\[4pt]

\textbf{Final translation:}
\textit{He is robbing small bells at the ear.}
&
\textbf{Final translation:}
\textit{Cover ears while stealing a bell.}
\\

\bottomrule
\end{tabularx}
\caption{Side-by-side comparison of the generated and gold reasoning traces for a Xibe-to-English translation example. Long grammar rules are omitted for space. Although the generated trace closely follows the required linguistic reasoning format, it contains many errors in lexical selection and syntactic analysis. Errors are annotated in \textcolor{errorred}{red}, and the corresponding correct analyses are marked in \textcolor{blue}{blue}.}
\label{tab:sjo-qualitative-example}
\end{table}

\end{document}